\documentclass[sigconf]{acmart}

\renewcommand\footnotetextcopyrightpermission[1]{} 
\makeatletter
\renewcommand\@formatdoi[1]{\ignorespaces}
\makeatother

\usepackage{times}
\usepackage{latexsym}
\usepackage{graphicx}  

\usepackage[T1]{fontenc}

\usepackage[utf8]{inputenc}

\usepackage{microtype}
\usepackage{array,multirow}
\usepackage{tikz}
\usepackage{booktabs} 
\usepackage{framed} 
\usepackage{enumitem}
\usepackage{subcaption}
\usepackage{tabulary}
\usepackage{pifont}
\newcommand{\cmark}{\ding{51}}%
\usepackage{adjustbox}
\newcolumntype{R}[2]{%
    >{\adjustbox{angle=#1,lap=\width-(#2)}\bgroup}%
    l%
    <{\egroup}%
}
\newcommand*\rot{\multicolumn{1}{R{45}{1em}}}

\newcommand{\acro}{{\sc Goal}}
\newcommand{\VideoSet}{\ensuremath{\mathcal{V}}}
\newcommand{\Video}{\ensuremath{v}}
\newcommand{\VideoChunk}[2]{\ensuremath{v_{#1,#2}}}

\newcommand{\VideoFeatures}{\ensuremath{\mathbf{V}}}
\newcommand{\VideoChunkFeatures}[2]{\ensuremath{\mathbf{V}_{#1,#2}}}

\newcommand{\Caption}[2]{\ensuremath{c_{#1,#2}}}

\usepackage{lipsum}
\AtBeginDocument{%
  \providecommand\BibTeX{{%
    \normalfont B\kern-0.5em{\scshape i\kern-0.25em b}\kern-0.8em\TeX}}}

\setcopyright{none}
\acmDOI{none}
\begin{document}

\title{Going for \acro: A Resource for Grounded Football Commentaries}

\author{Alessandro Suglia}
\affiliation{%
  \institution{Heriot-Watt University}
  \city{Edinburgh}
  \country{UK}}
\email{a.suglia@hw.ac.uk}

\author{José Lopes}
\affiliation{%
  \institution{Semasio, Lda.}
  \city{Porto}
  \country{Portugal}}
\email{jose@semasio.com}

\author{Emanuele Bastianelli}
\affiliation{%
  \institution{Heriot-Watt University}
  \city{Edinburgh}
  \country{UK}}
\email{emanuele.bastianelli@gmail.com}

\author{Andrea Vanzo}
\affiliation{%
  \institution{Heriot-Watt University}
  \city{Edinburgh}
  \country{UK}}
\email{andrea.vanzo1@gmail.com}

\author{Shubham Agarwal}
\affiliation{%
  \institution{Heriot-Watt University}
  \city{Edinburgh}
  \country{UK}}
\email{sa201@hw.ac.uk}

\author{Malvina Nikandrou}
\affiliation{%
  \institution{Heriot-Watt University}
  \city{Edinburgh}
  \country{UK}}
\email{mn2002@hw.ac.uk}

\author{Lu Yu}
\affiliation{%
  \institution{Heriot-Watt University}
  \city{Edinburgh}
  \country{UK}}
\email{l.yu@hw.ac.uk}

\author{Ioannis Konstas}
\affiliation{%
  \institution{Heriot-Watt University}
  \city{Edinburgh}
  \country{UK}}
\email{i.konstas@hw.ac.uk}

\author{Verena Rieser}
\affiliation{%
  \institution{Heriot-Watt University}
  \city{Edinburgh}
  \country{UK}}
\email{v.t.rieser@hw.ac.uk}

\renewcommand{\shortauthors}{Suglia, et al.}


\begin{abstract}
Recent video+language datasets cover domains where the interaction is highly structured, such as instructional videos, or where the interaction is scripted, such as TV shows. Both of these properties can lead to spurious cues to be exploited by  models rather than learning to ground language.
In this paper, we present GrOunded footbAlL commentaries (\acro), a novel dataset of football (or `soccer') highlights videos with transcribed live commentaries in English. As the course of a game is unpredictable, so are commentaries, which makes them a unique resource to investigate dynamic language grounding.
We also provide state-of-the-art baselines for the following tasks: frame reordering, moment retrieval, live commentary retrieval and play-by-play live commentary generation. Results show that SOTA models perform reasonably well in most tasks. We discuss the implications of these results and suggest new tasks for which \textsc{Goal} can be used. Our codebase is available at: \url{https://gitlab.com/grounded-sport-convai/goal-baselines}.
\end{abstract}

\begin{teaserfigure}
  \centering
  \includegraphics[width=0.65\textwidth, keepaspectratio]{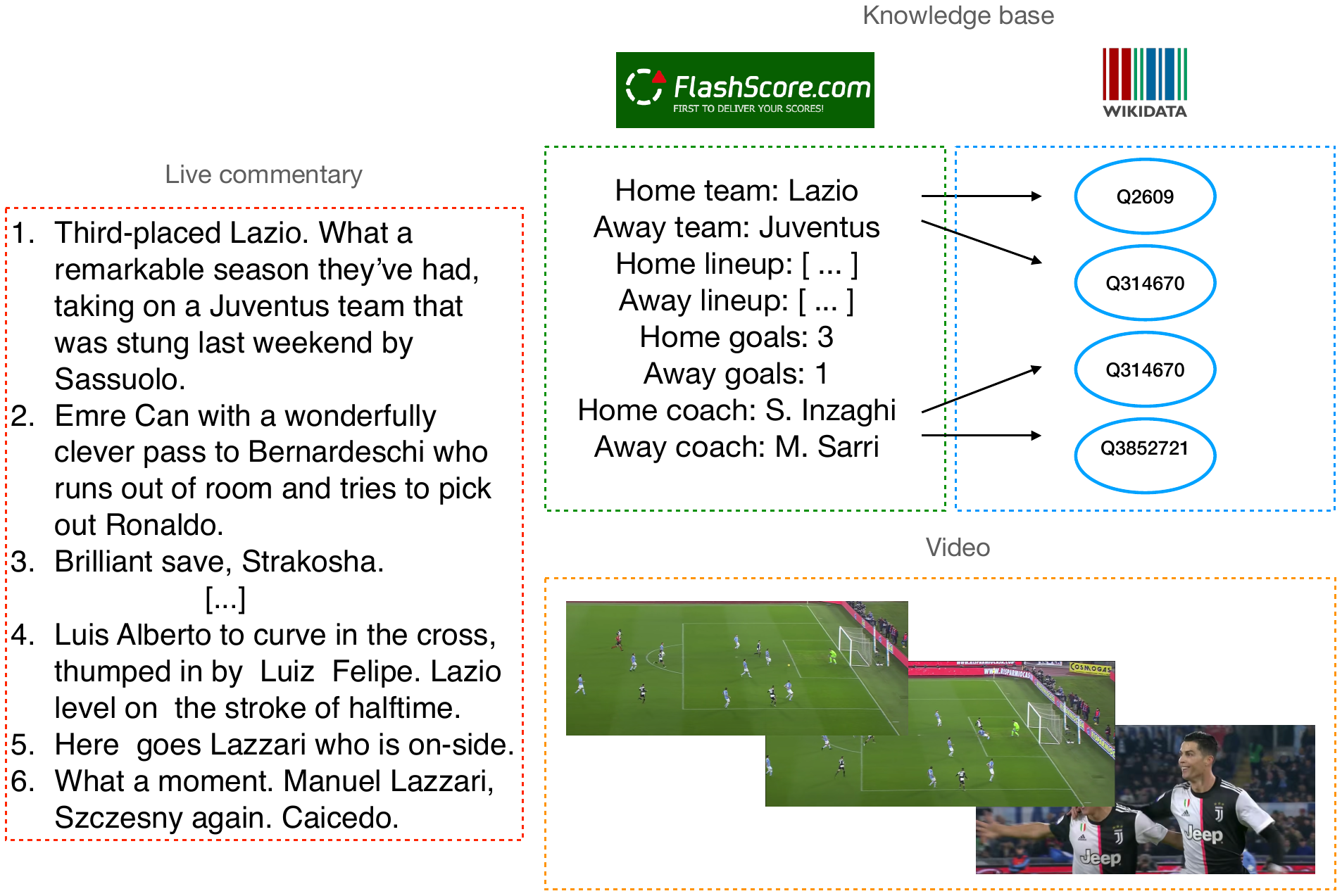}
  \caption{Overview of the resources that are part of the \acro{} dataset. We provide: videos from international football leagues associated with professionally transcribed commentary;  
  a knowledge base containing players, teams and their managers for every match; 
 entity grounding on Wikidata as further factual knowledge.}
 \label{fig:goal_overview}
\end{teaserfigure}

\maketitle

\pagestyle{plain}

\section{Introduction}
Research on visual language grounding and visual dialogue combines recent advances in computer vision and natural language processing by grounding conversation in image~\cite{shuster2020image} or video data~\cite{alamri2019audiovisual}. Whereas open domain dialogue systems seek to embody their own `persona' to be entertaining~\cite{see2019makes}, and some systems use external knowledge to reason~\cite{yang2020graphdialog}.
In this research, we propose the domain of sports as a test-bed which combines
several elements of multimodal and grounded natural language processing. 
In particular, we collect a new dataset on football commentaries (see transcript in Figure~\ref{fig:goal_overview}) which contains speech, transcribed text, real background audio and video data. The task requires visual grounding as in video captioning generation (see transcript 2,3,4,5,6), integrating external knowledge sources, such as knowledge about standings (1), personal opinion (2,3,6), previous events/history (4), while being entertaining (6). So far, none of the current state-of-the-art models is able to address all these challenges at once. For example, the recent multi-modal Transformer model~\cite{tsai2019multimodal} only combines audio, video and captions. Therefore, \acro{} represents a unique opportunity to develop and evaluate models with a larger number of input modalities.

As such, the \acro{} dataset aims to complement current efforts in providing large scale datasets for benchmarking knowledge and image grounded systems~\cite{shuster-etal-2020-dialogue} and video-subtitle moment retrieval~\cite{Lei2020tvr}. For example, \citet{shuster-etal-2020-dialogue} propose the Dialogue Dodecathlon, which assembles existing datasets representing various sub-tasks, including asking and answering questions, responding to statements, as well as persona, knowledge, situation and image grounding. 
In contrast to the Dodecathlon benchmark, our dataset requires these skills to be demonstrated within a single task, which we believe makes it a more challenging setup.

In addition, \acro{} aims to probe existing pretrained models in a setup which is closer to the one envisaged by \citet{de2020towards} by creating a corpus from live football commentaries, where video and language are not scripted but instead are collected from a realistic setting involving complex visual scenes, and articulated (noisy) language commentaries. 

As such, the main contributions of this paper are as follows:
1) we release \acro, a new dataset based on football commentaries that requires advances in several fields such as visual language grounding, data-to-text generation and multi-modal fusion;
2) we provide an evaluation framework based on frame reordering, commentary retrieval, commentary generation and, video moment retrieval;
3) we compare variants of the state-of-the-art video+language model presented in \cite{li-etal-2020-hero} on these sub-tasks and compare the results to uni-modal ablations.

\section{Background and Related Work}

\begin{table*}[]

    \centering
    \begin{tabular}{cccccccccc}
        Name & Domain & \rot{Situation Grounding} & \rot{KB  Grounding} & \rot{Manual Annotations} & \rot{Spoken} & \rot{Spontaneous} & \rot{Language} & \# Videos & Avg. Duration (s)\\
        \hline
        \textsc{Tacos} \cite{regneri-etal-2013-grounding} & Cooking & \cmark & & \cmark & & & EN & 127 & 260 \\ 
        YouCook II \cite{ZhXuCoAAAI18} & Cooking & \cmark & & \cmark & & & EN & 2,000 & 326 \\
        HowTo100M \cite{sanabria18how2} & Instructional & \cmark\textsuperscript{*} & & & \cmark & \cmark & EN/PT & 79,114 & 90\\
        MSR-VTT \cite{Xu2016MSRVTT} & Various & \cmark\textsuperscript{*} & & \cmark & & & EN & 10,000 & 10-30 \\
        ActivityNet Captions \cite{Kishna2017ActivityNet} & Various & \cmark & & \cmark & & & EN & 20,000 & 120 \\       
        DiDeMO \cite{hendricks-2017-didemo} & Various & \cmark\textsuperscript{*} & & \cmark & & & EN & 10,464 & 5\\
        ViTT \cite{huang-etal-2020-multimodal} & Various & \cmark & & \cmark & & & EN & 8,169 & 7.1 (segments) \\        
        TVR \cite{Lei2020tvr} & TV shows & \cmark\textsuperscript{*} & & \cmark & \cmark & & EN & 21,793 & 76.2 \\
        \textsc{Violin} \cite{liu2020violin} & TV-shows & \cmark & & \cmark & \cmark & & EN & 95,322 & 90 \\
        Twitch-FIFA \cite{pasunuru-bansal-2018-game} & Video-games & \cmark\textsuperscript{*} & & & & \cmark & EN & 49 & 1.76 (hours)\\
        \acro & Football Commentary & \cmark & \cmark & \cmark & \cmark & \cmark  & EN & 1,107 & 238 \\ 
        \hline
    \end{tabular}
    \caption{\acro{} vs existing video+language datasets, in terms of features, language and size. (\textsuperscript{*}) means that not all the language is grounded in the visual scene or in any piece of external knowledge (e.g. chit-chat in TV-shows).}
    \label{tab:dataset_comparison}
\end{table*}

Live sports events are broadcasted daily around the world. Broadcasts have journalists and commentators who provide descriptions of the events happening during the match, technical insights and facts about athletes taking part in the event. As commentators refer to the game they are watching, their language is intrinsically grounded on the events occurring  as well as on general knowledge about teams and players involved in the game. In addition, commentators rely on game statistics to improve the viewer experience both from an entertaining and technical perpective. Bringing all these dimensions together to generate human-like football commentary unites several on-going research challenges from knowledge grounded data-to-text generation, and several video+language tasks. Table~\ref{tab:dataset_comparison} provides a comparison between previous resources in these sub-fields and \acro{}'s contributions to the community.

\subsection{Data-to-text generation}
Data-to-text (D2T) generation takes structured data, such as RDF triples~\cite{gardent-etal-2017-creating} or tabular data~\cite{novikova2017e2e}, and translates it to natural language. 
D2T has been applied to sports statistics \cite{liang-etal-2009-learning,wiseman-etal-2017-challenges,barzilay-lapata-2005-collective,kelly-etal-2009-investigating,nie-etal-2018-operation,puduppully-etal-2019-data},
by using tabular data to generate an end-of-match summary report. 
\citet{van-der-lee-etal-2017-pass} propose a template-based football report generation method, which  
was further enhanced in \cite{gatti-etal-2018-template} by creating a link to Wikidata. 
\acro{} can be seen as an extension of these D2T generation tasks with the help of vision, language and audio features. 
Furthermore, we address live commentary, rather than end-of-match report generation, which has to take live events into account,  
similar to play-by-play commentary generation \cite{koncel-kedziorski-etal-2014-multi,Taniguchi_Feng_Takamura_Okumura_2019} and alignment to game events \cite{Hajishirzi2012Oopta}.
Recent work \citep{zhang2021soccer}, has contributed to this research by releasing a new dataset of play-by-play comments and events in more than 2,000 matches.
However, in contrast to play-by-play commentary, which is added post-hoc and as such is written and clean, \acro{} contains transcriptions of live commentary with all the properties of spontaneous speech, such as repairs or unfinished sentences. 
Live commentary generation for sports broadcast has been previously investigated by \citet{ChenSportcast2008} but for simulated football matches only, where a structured representation of the visual scene is 
 available.
 
\subsection{Video+language datasets}

In contrast to D2T approaches, which `ground' in external facts, current video+language datasets mostly focus on situational grounding. 
Initial approaches to video+language, such as \citet{Karpathy2014Sports100M}, cast video understanding as a multi-class classification problem where labels
are based on a class taxonomy for sports activities.  
A parallel line of research has focused on grounded instructions, such as cooking
\cite{regneri-etal-2013-grounding,ZhXuCoAAAI18}, where datasets contain time aligned textual description of events for downstream task for caption generation \cite{Zhou2018Dcap} and retrieval \cite{luo2020univl}.
One of the limitations of instructional videos is their highly structured nature, which allows models to learn task cues rather than ground language.
In contrast, datasets such as ActivityNet \cite{Kishna2017ActivityNet}, DiDeMo \cite{hendricks-2017-didemo} and MSR-VTT \cite{Xu2016MSRVTT} contain randomly selected content. For example, MSR-VTT contains videos ranging from sports to political speeches.
Lastly, 
datasets created from TV series and movies, such as  LSMDC \cite{lsmdc2015}, VIOLIN \cite{liu2020violin} or TVR \cite{Lei2020tvr}, also span a wide variety of domains and situations, but are highly scripted due to the dialogical nature of these TV productions.

In contrast, sports videos are not structured as instructional videos, nor are they scripted as TV shows or movies. The action is shot from different angles and often replays are used. 
Recently, football datasets have received increased interest in the vision community for 
action recognition
\cite{Karpathy2014Sports100M,deliege2020soccernetv2,Cioppa_2020_CVPR}.
The closest to the work presented here is
\citet{pasunuru-bansal-2018-game}. However, in contrast to \acro, \citeauthor{pasunuru-bansal-2018-game} do not aim to generate professional commentary, but written
chit-chat between spectators watching games on Twitch.tv. 
The authors report that more than half of the chat is not related to the video and thus not useful to evaluate grounded models.

\section{GOAL Benchmark}

\subsection{Dataset preparation}

\acro{} contains 1,107 video highlights from football matches from major competitions in Europe between 2018 and 2020 available on YouTube.
We selected the videos so that 1) they only contained highlights 2) the commentaries are in English\footnote{We are planning to extend the dataset to other languages in a future release.} and 3) they have a minimum duration of $80$ seconds.
Following other dataset creation initiatives (e.g., ~\cite{lei-etal-2020-tvqa,ZhXuCoAAAI18,lee2021acav100m}), we will publish the data following fair use. Specifically, interested researchers, after signing an NDA, will have access to the source URLs, transcriptions produced by professional transcribers,
information about players, teams, line-ups and matches, extracted video features~\cite{li-etal-2020-hero} and audio features extracted with OpenSmile using the IS2010 feature set \cite{Eyben2010OpenSmile}. 

\subsubsection{Transcriptions curation}

We use professional transcription services~\footnote{We used \href{https://gotranscript.com/}{GoTranscript} for our data collection.} rather than Automatic Speech Recognition  due to the specific sub-register used at sports events (e.g.\ specific terminology as well as player and team names) 
and due to the presence of
background noise.
In order to support the transcription process, we provided the transcribers with the knowledge base information for a specific match (e.g. line-ups, substitutions, ...). Transcriptions were provided as captions paired with the corresponding timestamps (i.e. start and end). For consistency, sections which were considered unintelligible by the professional transcribers were manually verified. The majority was resolved, leaving only a few still unintelligible, marked with the token \texttt{[?]}.

\begin{table*}[tb]
\centering
\begin{tabular}{|lll|lll|lll|}
\hline
\multicolumn{3}{|c|}{Train}          & \multicolumn{3}{c|}{Validation}      & \multicolumn{3}{c|}{Test}            \\ \hline
\# Scenes & \# Sentences & \# Tokens & \# Scenes & \# Sentences & \# Tokens & \# Scenes & \# Sentences & \# Tokens \\
6623      & 29295        & 10815     & 1728      & 7608         & 5700      & 3520      & 16123        & 8186      \\ \hline
\end{tabular}
\caption{GOAL dataset statistics reported for the training, validation, and test splits generated using stratified sampling based on the football league of each video.}
\label{tab:goal_dataset_stats}
\end{table*}

Transcriptions are thus grouped by start and end timestamps identifying the time frame in which a specific caption chunk has been uttered. However, sentences referring to a specific event may be split in several continuous chunks, breaking the syntax and meaning of the whole sentence itself.
For example, chunk 2 in Figure \ref{fig:goal_overview} may have been split into two different chunks due to the timestamp-based grouping, with the first chunk \textit{``Emre Can with a wonderfully clever pass to Bernardeschi''} ending at time $t_i$ and the second one \textit{``who runs out of room and tries to pick out Ronaldo.''} starting at time $t_j$ (with $t_j > t_i$). Similarly, chunks 5 and 6 clearly refer to the same event even if they have been split.

In order to perform the tasks designed in this benchmark, complete and well-formed sentences had to be reconstructed, i.e. all chunks composing a sentence or referring to the same event had to be merged. To satisfy these requirements, we defined two alternative heuristics to refine and post-process the output transcriptions:
\begin{enumerate}
    \item \textit{sentence-based} -- two consecutive chunks $s_t$ and $s_{t+1}$ are merged into a single chunk whenever $s_t$ does not end with an \textit{end-of-sentence} punctuation (e.g. full stop). This heuristics would allow to merge chunk 2 of Figure \ref{fig:goal_overview} if they were originally split, due to the first sentence not ending with a period.
    \item \textit{event-based} -- two consecutive chunks $s_t$ and $s_{t+1}$ are merged into a single chunk if the distance between the end timestamp of $s_t$ and the start timestamp of $s_{t+1}$ is lower than a $2.5$. This heuristics would merge chunks 5 and 6, if the difference between start timestamp of 6 and end timestamp of 5 is $< 2.5$.
\end{enumerate}

Using these post-processing heuristics, we obtain two different versions of the dataset; specific versions of the dataset are used by specific tasks in the \acro{} benchmark depending on their temporal granularity. Please refer to Section~\ref{sec:exp_eval} for details.

\subsubsection{Dataset statistics}

The resulting dataset has been divided into 3 splits using stratified sampling based on the league label. We completed a first 70-30 sampling obtaining \texttt{train+val} and \texttt{test} splits. Then, we completed a 80-20 sampling obtaining the \texttt{train} and \texttt{val} splits. These splits are the ones that we use as reference for all the tasks that are presented in the \acro{} experimental evaluation. First of all, in Table~\ref{tab:goal_dataset_stats} we report the statistics about the reference splits that are used in all the proposed tasks. In order to demonstrate the complexity of the language contained in \acro{}, we used the spaCy toolkit~\footnote{\url{https://spacy.io/}}, to calculate the vocabulary size \emph{without} named entities. To do that, we extracted all the sentences in the dataset and for each of them, we extracted all the unique tokens associated with named entities. We removed them from our vocabulary giving us a vocabulary of size $6819$ tokens for the training set. This gives us an idea that a delexicalisation process will not overly simplify the language complexity of the dataset, classifying the \acro{} dataset as challenging dataset for the Vision and Language community. The $1107$ videos are extracted from $8$ different leagues such as: the Italian Serie A ($638$ videos), the English Premier League ($167$ videos), UEFA Champions League ($62$ videos), UEFA Europa League ($61$ videos), FA Cup ($34$ videos), Carabao Cup ($18$ videos), EFL Championship ($77$ videos) and EURO 2020 Qualifiers ($50$ videos).

\subsubsection{Knowledge base integration}
Football commentators often resort to their knowledge of the football scene, as well as to additional external sources about players, line-ups, teams and so on. We construct a knowledge base to surrogate such factual information, so that it can be exploited to extend the capability of a model.
We extract all the information about line-ups, play-by-play commentary, summaries of game events, teams, as well as personal data about players (e.g. age, position, nicknames), from \url{www.flashscore.com} for each of the 1,107 matches.
In addition, we link teams and players to the corresponding Wikidata entity~\cite{vrandevcic2014wikidata} by retrieving their Wikidata identifier, to augment the knowledge base with semantic information following Linked Open Data principles~\cite{bizer2011linked}. 

\subsection{Tasks Description}\label{ssec:taskdescr}

To demonstrate the utility of the dataset, we have conducted an evaluation on four tasks that involve both grounded natural language understanding and generation skills following \cite{li-etal-2020-hero}: 1) \textit{commentary retrieval}, 2) \textit{frame reordering}, 3) \textit{video moment retrieval}, and 4) \textit{commentary generation}. 

\paragraph{Notation:}
Every video \Video{} in the set of videos \VideoSet{} has been divided into $K$ chunks using the heuristics described above. Every chunk \VideoChunk{s}{e}$\in v$ is identified by start and end timestamps ($s$ and $e$, respectively), and paired to the caption \Caption{s}{e} of the corresponding live commentary between timestamp $s$ and $e$. 

\paragraph{Commentary retrieval:} 
We formulate this task akin to response retrieval tasks as used for 
image-based \cite{das2017visual,shuster2020image, moon2020situated} as well as video-based dialogues \cite{alamri2019audiovisual}. We use the \textit{event-based} version of our dataset for this task. In particular, given a chunk of the video \VideoChunk{s}{e} as well as previous history $H_{v_{s,e}}=\{v_{*,p} : v_{*,p} \in v \land p \leq s\}$, a model has to select the most appropriate candidate response $\tilde{r}$ among a set of $n$ candidates $\mathcal{R}_{\VideoChunk{s}{e}}$.
In this setup, we assume that $\mathcal{R}_{\VideoChunk{s}{e}}$ is composed of a ground-truth (GT) response $\hat{r}$ and $n-1$ negative candidates. The GT corresponds to the caption of the video chunk $\Caption{s}{e}$.  
The negative candidates are sampled uniformly at random from the entire dataset. The average sentence similarity between the ground truth and the negative candidates was $0.35$ in the training set\footnote{Sentence similarity computed using the \texttt{stsb-roberta-large} model available in \url{https://github.com/UKPLab/sentence-transformers}. We use the cosine similarity to compare the sentence embeddings.}. Future work will explore more sophisticated techniques for adversarial candidate generation \cite{zellers2019recognition,le2020adversarial}. To evaluate the performance of models in this task, we use retrieval-based metrics similar to~\cite{das2017visual,alamri2019audiovisual,shuster2020image, moon2020situated}: 1) \textit{R@1}: GT response is in first position of the ranked list (higher is better); 2) Mean Reciprocal Rank (\textit{MRR}): 
average of the reciprocal position
of the GT response (higher is better); 3) Mean rank (\textit{mean}): average position of the GT response (lower is better). For this evaluation, we limit the number of candidates $n$ to 5.


\paragraph{Frame reordering:} 
Similar to the pretraining task in \texttt{HERO} \cite{li-etal-2020-hero},
we define a frame reordering task for the \textit{event-based} version of the dataset. 
This task serves to understand whether the model is able to learn the event dynamics of the video. This is especially relevant since
in general, football videos have a faster pace than cooking videos or TV shows. 
We divide a given video chunk \VideoChunk{s}{e} into $k$ sequential video features $\VideoChunkFeatures{s}{e} \in \mathbb{R}^{k \times h}$.
We shuffle $15\%$ of them and ask the model to predict the correct frame indexes. We formulate this as a classification task where the model has to predict, for every visual feature, a label in the set $\{1, 2, \dots, M\}$ where $M$ is the maximum number of visual features that can belong to a given video, in our case $100$. Therefore, the corresponding label setting ranges from 0 to 99. We evaluate model performance based on categorical accuracy. 

\paragraph{Moment retrieval:} This task follows the \textit{Video-Subtitle Matching} pretraining task presented in \cite{li-etal-2020-hero} and uses the \textit{sentence-based} version of the dataset. 
Given 1) $m$ video features $\VideoFeatures{} \in \mathbb{R}^{m \times h_v}$ (where $h_v$ is their dimensionality), and 2) a caption \Caption{s}{e}, the model has to predict the start and end indexes ($\tilde{s}$, $\tilde{e}$) of \Caption{s}{e}~\footnote{We automatically derive the alignment between visual features and captions based on timing overlap.}.
It is worth noting that $\tilde{s}$ and $\tilde{e}$ are predicted independently; this may give rise to invalid combinations (i.e. $\tilde{s} > \tilde{e}$).
To deal with invalid spans, at inference time we select the span with the maximum likelihood among the valid spans only (i.e. $s < e$) . We evaluate the performance of the models 
based on two metrics: 1) \texttt{Soft}: average number of spans that have a \textit{soft} match, i.e.\ 
whenever there is a partial overlap between the predicted span $(\tilde{s}, \tilde{e})$ and the gold ones $(s, e)$; 2) \texttt{Weighted}: 
similar to Intersection over Union which evaluates the proportion of matching predicted indexes compared to the gold ones.

\paragraph{Commentary generation:}
The intuition for this task is the following: imagine watching a short video snippet from timestamp $s$ to $e$, pausing and then generating a description for that chunk. 
Note that a given caption for a chunk \Caption{s}{e} can contain several $L$ sentences $\{\Caption{s}{e}^1, \Caption{s}{e}^2, \dots, \Caption{s}{e}^L\}$. 
The model has to generate one sentence at a time conditioned on both the video and on what has been generated so far (i.e. the commentary history). 
For this task we use the \textit{sentence-based} version of GOAL and report standard NLG evaluation metrics, including \textsc{BERT}score~\cite{zhang2019bertscore}, \textsc{Bleu@4} \cite{papineni2002bleu}, \textsc{Meteor} \cite{lavie2007meteor} and \textsc{Rouge-L} \cite{lin2004automatic}. We also perform a detailed error analysis to cope with the fact that current NLG metrics are not well correlated with human judgements~\citep{novikova2017we}.

\section{Experimental evaluation} \label{sec:exp_eval}

This section reports the results for each of the tasks outlined in Section \ref{ssec:taskdescr}, where we compare 
a fine-tuned \texttt{HERO} model 
\cite{li-etal-2020-hero}\footnote{Due to computational constraints, we were able to run the HERO-flat version of the model only.} to uni-modal baselines. We decided to use \texttt{HERO} because it's a large-scale Video+Language (V+L) model considered the baseline model for \textit{VALUE}, another recent V+L benchmark~
\cite{li2021value}. This comparison aims to assess the extent to which fusing the two modalities is required to complete the task~\cite{thomason2019shifting}. We used the \acro{} validation set for hyperparameter tuning and the test set as a benchmark for the evaluation reported in the paper. Please refer to Appendix for details about the selected hyperparameters. We are planning to release the codebase that enables users to reproduce our experiments.

\subsection{Commentary retrieval}
In this task, we assessed the ability of the model to select the best response among a set of candidates, similar to the Visual Dialog (VisDial) task specified in \citet{das2017visual}. We are interested in the following questions: 1) are all modalities useful to complete the task? 2) is \emph{incorporating} and \emph{using} the commentary history important to do well? The first question is motivated by the fact that datasets like VisDial and VQA often can be `solved' by exploiting language artifacts.
The latter is motivated by \citet{agarwal2020history}'s findings that the VisDial only requires minimal history context.
To answer the first question, we compared the \texttt{HERO} model with two uni-modal baselines: 1) Vision-only \texttt{HERO}: a \texttt{HERO} model that  is only using the current video to select the candidate; 2) \texttt{RoBERTa}: a RoBERTa model fine-tuned on the \acro{} dataset using a multiple choice objective~\cite{liu2019roberta}. In addition, to assess the importance of encoding the history, 
we evaluated {\tt HERO} with different number of textual tokens as part of the history (h) as well as different number of captions (c) (where $h$ represents an upper bound on the number of tokens included).

Results in Table~\ref{tab:retrieval_results} show that a strong text-only baseline like \texttt{RoBERTa} performs marginally above chance -- 
indicating that video is required 
to correctly select the next response. 
The \texttt{HERO} (no history) model, which encodes visual features and the candidate response using the \texttt{HERO} Cross-modal Transformer, shows an increase of $+19.4\%$ in R@1, which confirms that vision is useful.
However, this model does not take into account the history. 
By experimenting with different levels of complexity for history encoding,  we observe that encoding just the last $50$ tokens ($h=50$) of the previous caption ($c=1$) gives a boost in performance of $+1.6\%$ over the no history model. 
By further increasing the history all retrieval metrics continue improving.
In sum, our results show that models trained on \acro\ require both multi-modal fusion and history, which makes this dataset better suited to study grounded language than datasets such as VisDial or VQA.

\begin{table}[]
\centering
\begin{tabular}{@{}llll@{}}
\toprule
                  & \textsc{R@1}   & \textsc{MRR}   & \textsc{Mean}  \\ \midrule
\texttt{RoBERTa} (h=50, c=1)    & $36.3\%$ & 0.593 & 2.320 \\
\texttt{HERO} (no history)  & $55.7\%$ & 0.737 & 1.746 \\
\texttt{HERO} (h=50, c=1)  & $57.3\%$ & 0.748 & 1.715 \\
\texttt{HERO} (h=50, c=2)  & $58\%$ & 0.755 & 1.673 \\
\texttt{HERO} (h=80, c=2)  & $59.9\%$ & 0.765 & 1.648 \\
\texttt{HERO} (h=100, c=2) & $63\%$  & 0.79  & 1.58  \\ \bottomrule
\end{tabular}
\caption{Results of the response retrieval evaluation on the event-based GOAL dataset. We compare \texttt{HERO} with uni-modal baselines to understand the utility of using both modalities for the task. We also experiment with different numbers of history tokens (h) and different numbers of captions (c) that compose the history. }
\label{tab:retrieval_results}
\end{table}

\subsection{Frame reordering}

We intend \textit{frame reordering} as a diagnostic task to assess whether multi-modal models retain the ability to infer the order of the encoded visual inputs.
We compare \texttt{HERO} with several baselines using the same pretrained visual features. 
 In particular, we report results for four baseline sequence encoders that use an MLP with two hidden layers to complete the classification task: 1) \texttt{MLP}: receives in input the visual features as input without positional information; 2) \texttt{MLP+pos}: 
 extends the previous model with positional embeddings; 3) \texttt{bi-GRU}: a one-layer bidirectional GRU~\cite{cho2014properties} generating contextual representation for the visual features; 4) \texttt{bi-LSTM}: same as before but using LSTM cells~\cite{hochreiter1997long}. Results are shown in Table~\ref{tab:frame_reordering_results}. We see that a model that does not use positional information (\texttt{MLP}) barely goes above $10\%$ 
 indicating that the task indeed requires positional information. When using expressive models that innately have a bias for sequential modelling such as LSTMs or GRUs, accuracy is almost perfect. However, \texttt{HERO} seems to
 sacrifice its ability to recover the original frame order to better encode other elements of its input. This result indicates the need for 
 a more systematic way of incorporating multi-modal information in single-stream transformer models as well as better ways of dealing with positional information across multiple modalities. 

\begin{table}[ht]
\centering
\begin{tabular}{@{}ll@{}}
\toprule
          & \textsc{Accuracy} \\ \midrule
\texttt{MLP}       & $10.42\%$  \\
\texttt{MLP+pos}   & $10.99\%$  \\ \midrule
\texttt{bi-GRU}       & $90\%$ \\
\texttt{bi-LSTM}       & $91\%$ \\ \midrule
\texttt{HERO} & $87\%$     \\ \bottomrule

\end{tabular}
\caption{Results for the frame reordering task for the \acro{} dataset.
For every instance, $15\%$ of the video frame features are shuffled.}
\label{tab:frame_reordering_results}
\end{table}

\subsection{Video moment retrieval}

\begin{table}[ht]
\centering
\begin{tabular}{@{}lll@{}}
\toprule
         & \textsc{Soft}           & \textsc{Weighted}       \\ \midrule
\texttt{HERO}     & $70.6\%$ & $2.6\%$ \\
\texttt{VRoBERTa} & $32.24\%$          & $2.11\%$          \\ \bottomrule
\end{tabular}
\caption{Results from the moment retrieval task for the \acro\ dataset. We compare the \texttt{HERO} model with a strong vision+language baseline \texttt{VRoBERTa}.}
\label{tab:results_moment_retrieval}
\end{table}

For video moment retrieval, we compare the \texttt{HERO} model with a late fusion baseline which we call \texttt{VRoBERTa}. 
{\tt VRoBERTa} concatenates 
visual embeddings (the same used in \texttt{HERO}) with the \texttt{[CLS]} hidden state representation generated by RoBERTa 
for the current caption. The resulting representation is passed through a feed-forward neural network to predict the likelihood of that frame being the start or end index. We keep the layout of the feed-forward network consistent between the two models for fair comparison. 

The results in Table~\ref{tab:results_moment_retrieval} show 
that cross-modal attention learned by \texttt{HERO} 
boosts performance by $+38.36$ and $+0.49$ for \textsc{Soft} and \textsc{Weighted} scores, respectively. 
An example-based analysis shows that predicted spans are fairly wide and therefore exact match scores are low. A possible explanation is the noise introduced by the automatic video features-subtitle alignment (cf. Appendix). 
Nevertheless, \texttt{HERO} achieves reasonable performance; we believe this is associated with \textit{video-subtitle matching} being a pretraining task for \texttt{HERO}~\cite{li-etal-2020-hero}.

\subsection{Commentary generation}

\subsubsection{Automatic Evaluation}

\begin{table}[]
\centering
\begin{tabular}{@{}lcccc@{}}
\toprule
        & \multicolumn{1}{l}{\textsc{BERT}} & \multicolumn{1}{l}{\textsc{BLEU}} & \multicolumn{1}{l}{\textsc{METEOR}} & \multicolumn{1}{l}{\textsc{ROUGE-L}} \\ \midrule
\texttt{CopyNet} &  $0.848$  & $0.41\%$  & $4.30\%$  & $9.05\%$\\
\texttt{BART} & $0.848$ & $0.79\%$ & $3.80\%$ & $8.79\%$ \\
\texttt{BART-kb} & $0.848$ & $0.84\%$ & $3.92\%$ & $8.78\%$   \\
\texttt{BART-kb-t}\textsuperscript{*} & $0.874$ & $2.28\%$ & $8.21\%$ & $21.2\%$ \\
\texttt{RoBERTa} & $0.845$ & $0.36\%$  & $2.94\%$ & $5.78\%$ \\
\texttt{HERO-t} & $0.847$                    & $0.53\%$                    & $3.55\%$                      & $6.69\%$                       \\
\texttt{HERO-nt} & $0.847$                    & $0.60\%$                    & $3.7\%$                      & $7.2\%$                       \\ \bottomrule
\end{tabular}
\caption{Results for commentary generation for the GOAL test set. We compare text-only decoders with different ablations of the \texttt{HERO} model. \texttt{BART-kb-t}\textsuperscript{*} is an upper bound oracle, since we used the target output to generate the KB triples.}
\label{tab:results_generation}
\end{table}

We train several text-only and vision-based models as commentary generation baselines. As text-only baselines we report \texttt{CopyNet}~\cite{gu2016incorporating}, a sequence-to-sequence model with a copy mechanism, \texttt{BART}~\cite{lewis2020bart}, an auto-regressive encoder-decoder model and \texttt{RoBERTa}, an encoder-decoder architecture 
initialised with \texttt{RoBERTa}'s weights 
for the encoder and decoder components following \cite{rothe2020leveraging}. We follow the same approach 
for our vision-based encoder-decoder architecture based on \texttt{HERO}. 
We experimented with two variants to investigate whether sharing cross-attention layers of the pretrained \texttt{HERO} encoder 
benefits the decoder training: 1) \texttt{HERO-t}: we tie the weights of the encoder and decoder allowing for complete parameter sharing;
2) \texttt{HERO-nt}: 
we only tie the word embedding layers of the \texttt{HERO} encoder and decoder, and fine-tune the decoder on the \textit{sentence-based} version of the \acro{} dataset.

To assess the usefulness of the KB during commentary generation, as showed in Table~\ref{table:hyperparam-test-kb}, we tested a \texttt{BART} variant where named entities are extracted from textual information in the commentary and used to retrieve information from the KB. This was done both from history (\texttt{BART-kb}) and from the target sentence to set an upper bound oracle (\texttt{BART-kb-t}\textsuperscript{*}). The retrieved KB information is represented in natural language and concatenated to the history. 
For example, if `team B' is identified in the context, the KB information is encoded 
as e.g.\ `team B is the away team'. Similarly for a `player A', we get `Player A is a team B striker'. We expect that by providing player positions to the model it will learn game related knowledge, such as 
a striker is more likely to score goals than a goalkeeper. We also experiment with the position in which the KB information is integrated. From our analysis, it seems that the model is always able to refer to the KB information no matter where it is located in the input.

\begin{table*}[h]
\centering
\begin{tabular}{lcc|cccc}
\multicolumn{7}{c}{\textbf{Commentary generation with KB priming}}\\
\hline
\textbf{Model} & \textbf{KB-source} & \textbf{PiC} & \textbf{BERTScore $\uparrow$} & \textbf{BLEU $\uparrow$} & \textbf{METEOR $\uparrow$} & \textbf{ROUGE-L $\uparrow$}  \\
\hline \hline
\texttt{BART} & history & first & 0.85 & 1.06 & 4.04 & 8.89 \\
\texttt{BART} & target & first & 0.85 & 1.25 & 4.34 & 9.82 \\
\texttt{BART} & history & last & 0.85 & 1.03 & 4.14 & 8.97\\
\texttt{BART} & target & last & 0.87 & 2.56 & 8.32 & 21.2 \\
\hline
\end{tabular}
\caption{Hyper-parameter tuning for \texttt{BART}-based models extended with the GOAL knowledge base. We report different configurations depending on the reference text used to extract the information from the GOAL KB (KB-source), and the position in the input text where the KB information is added (PiC); this can be either before the history (first) or after (last).}
\label{table:hyperparam-test-kb}
\end{table*}

\begin{figure*}[tb]
\centering
    \includegraphics[width=0.75\linewidth,keepaspectratio]{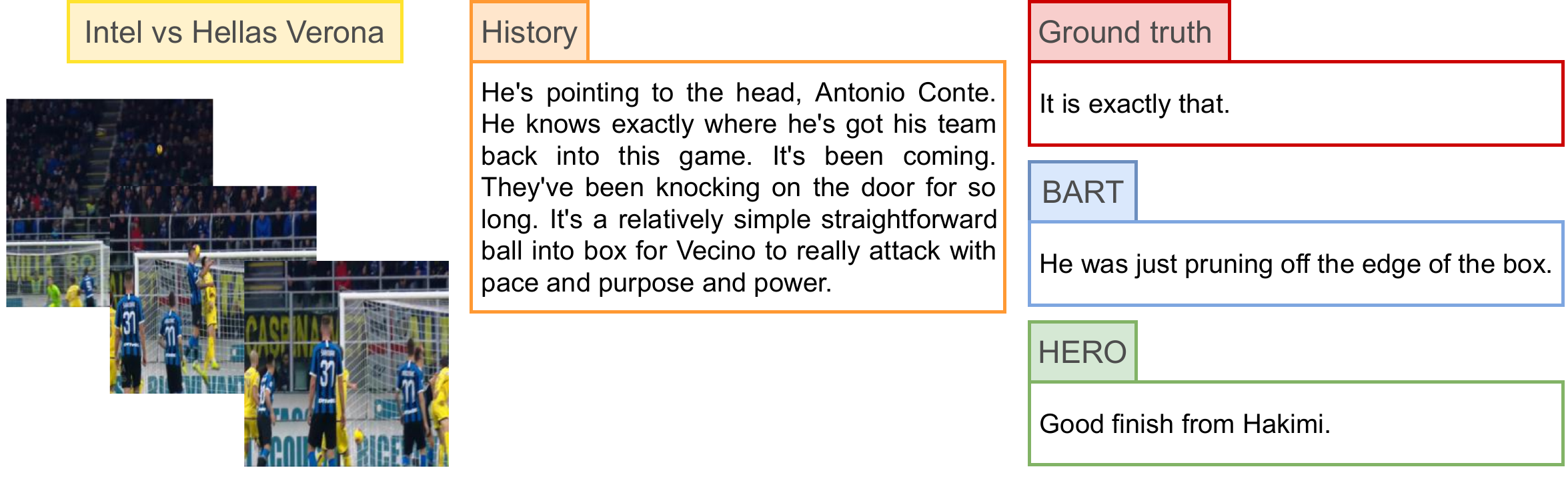}
  \caption{Example responses generated by the trained models for the commentary generation task. \texttt{HERO} uses both the video frames features and the history information to generate the response while \texttt{BART} only uses the textual history generated so far. This example is taken from a goal replay event.}
  \label{fig:generation_results}
\end{figure*}

All models 
use a history 
of $5$ previous sentences. We generate sequences token-by-token using nucleus sampling~\cite{holtzman2019curious} ($p=0.9$ for \texttt{CopyNet}, $p=0.95$ for the others) and top-k sampling~\cite{fan2018hierarchical} ($k=10$ for \texttt{CopyNet}, $k=100$ for the others). We truncate the generated output to $25$ tokens. 
The results in Table~\ref{tab:results_generation} show that all models perform relatively poorly. 
Even though \texttt{CopyNet} scores high on recall-based metrics METEOR and ROUGE, 
the lower BLEU score indicates less
coherent text compared to the pretrained baselines. 
Among those, \texttt{BART} outperforms the other text-only baseline \texttt{RoBERTa} by $0.43$ BLEU score, even though they are pretrained on the same data. When KB information is added, BLEU improves only by $0.05$. Despite  \texttt{HERO} and \texttt{RoBERTa} sharing the same encoder-only architecture,  \texttt{HERO}'s visual component enables 
better representation learning 
 with an gain of $+0.24$ BLEU score. Finally, 
 sharing all encoder-decoder layers (\texttt{HERO-t}) is not beneficial for this task. A more in-depth analysis across datasets is required to find 
 effective strategies for multi-modal encoder-decoder architectures.

\begin{table*}[]
\centering
\begin{tabular}{@{}lllllll@{}}
\toprule
     & \multicolumn{3}{c}{Language}      & \multicolumn{3}{c}{Language+Video} \\ 
     \cmidrule(l){2-7} 
     & Plausible & Repetition & Incoherent & Plausible  & Repetition & Incoherent \\ 
    \midrule
\texttt{BART} & 71\%      & 9\%        & 20\%       & N/A        & N/A        & N/A        
\\
\texttt{HERO} & 77\%      & 0          & 23\%       & 43\%       & 0          & 57\%       
\\ \bottomrule
\end{tabular}
\caption{Detailed results of the manual error analysis conducted on the output of the commentary generation models. The results are based on a sample of 100 contexts for which we manually analysed \texttt{HERO} and \texttt{BART} response predictions based on the defined annotation scheme.  We evaluate the generated commentary based on the ground-truth language or both language and video contexts.}
\label{tab:results_gen_annotations}
\end{table*}


\subsubsection{Error analysis}

We further investigate generation performance in an  error analysis since
automatic metrics 
do not fully capture model performance for language generation tasks, e.g.\ \cite{novikova2017we,kilickaya-etal-2017-evaluating}. 
We manually analysed 100 randomly selected generated outputs for the best performing models \texttt{HERO} and \texttt{BART}, 
using 3 labels to indicate performance levels:
\textit{plausible} caption, \textit{incoherent} text and \textit{repetition} of previous text.\footnote{Some generated examples can be found in Appendix.} 
For text-based models, our judgement was 
based only on the textual context given to the model (i.e.\ history). This means that a commentary suggesting ``the ball hit the crossbar'' is labelled as plausible given the goal chance context, even if the video was showing a goal. For the multi-modal model, we considered 
both textual and visual contexts. 

Results in Table~\ref{tab:results_gen_annotations} 
show that  \texttt{HERO} avoids repetition 
and produces mostly plausible outputs.
Indicating that  \texttt{HERO}'s visual features provide some innate bias towards more plausible responses influenced by representations that can capture high-level features of events (e.g. goals, penalty, etc.). This underlines our previous point that \acro\ requires both language and visual grounding in contrast to other Vision+Language tasks.
In particular, responses that are considered plausible from the text-only perspective, immediately become incoherent when observing the video. 
However, 
 \texttt{HERO}'s visual features are not 
fine-grained enough to capture details of the events, leading to a higher percentage of incoherent responses.

While the current improvement from the KB is small (\texttt{BART-kb}), we hypothesise that 
it could be more useful with entity extraction from vision (rather than the text history), e.g.\ by tracking the player who has the ball, using OCR to get their jersey number
and linking this information to the KB.
This hypothesis was tested with a simple experiment by extracting entities from the target sentence (\texttt{BART-kb-t}\textsuperscript{*}) boosting the performance of the model. 
In future work, we aim to 1) improve the KB representation during the generation process 2) share global match information with the highlight videos (e.g. play-by-play commentary) and 3) use the audio features to help identify certain events (e.g. goals from crowd cheering). 

\section{Situated Language Analysis}

Language in football commentaries is notoriously event-driven. Commentators describe how the game unfolds and how players interact, making football commentaries an interesting benchmark for learning grounded verb semantics from video. To understand how linguistic predicates relate to visual events in the dataset, we 
use the output of the state of the art semantic role labeller \textit{InVeRo}~\footnote{\url{http://nlp.uniroma1.it/invero/}} \cite{conia2020invero} as silver annotations
(which has high accuracy in both predicate and argument prediction of ~86.1\% and ~84\%, respectively). The objective of this annotation step is to have a sense of how ``situated" the collected data is. Specifically, we are interested in how predicative structures of event verbs describe actual situations happening in videos, providing evidence of the correspondence between the commentary and visual streams
(e.g., “player A \textit{passes} to player B”), and posing a solid base for language grounding. For a more detailed analysis, please refer to the Appendix.

\begin{figure*}[tb]
\centering
    \includegraphics[keepaspectratio, width=0.7\textwidth]{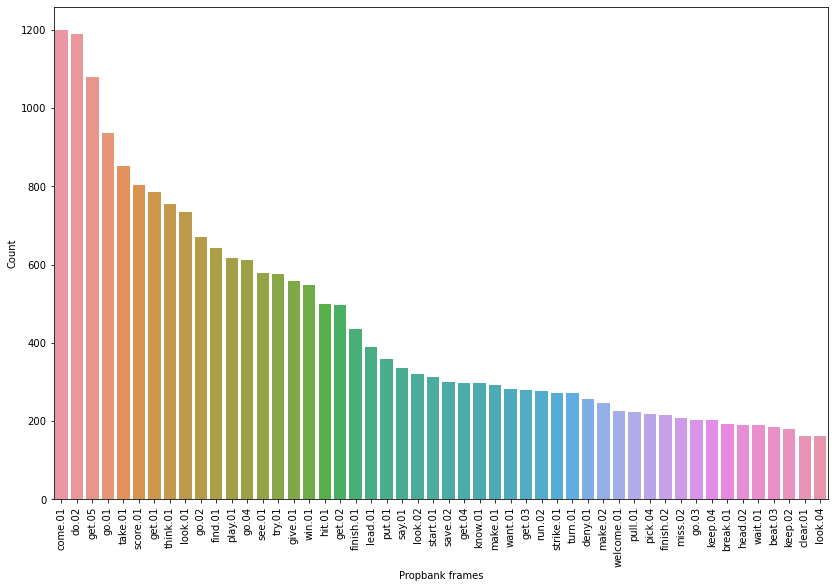}
  \caption{Distribution of the most common English Propbank predicates extracted from \acro{}, after removing predicates with modal and auxiliary function.}
  \label{fig:goal_srl}
\end{figure*}

Thanks to this automatic procedure, 
we extracted $68559$ predicative structures from \acro{} commentaries, ignoring sentences without verbs. 
As we can see from Figure~\ref{fig:goal_srl}, common verbs like \textit{do} and \textit{get} very often get the senses \texttt{do.02} and \texttt{get.05}, which represent physical actions and movements, relating to events happening in the videos. Verbs with high relevance to football jargon appear often in their ``sportive" sense, such as \texttt{score.01} and \texttt{strike.01}, and also more ambiguous verbs assume a more action-related sense, e.g., \textit{miss}, whose sense \texttt{miss.02} describes a \textit{no hit} action. Other verbs relating to the progress and the status of the match are also often used, such as \textit{play}, \textit{win}, \textit{lead}, and \textit{beat}. Such verbs, appear in their sport declination, i.e., \texttt{play.01}, \texttt{win.01}, \texttt{lead.01}, and \texttt{beat.03}, respectively. All these occurrences of verb senses show how the language used in the commentaries actually refers to events being displayed in the videos, thus providing evidence of the situated nature of the dataset. An interesting case is the one of the verb \textit{clear}. This verb is often used in commentaries with the sense of ``clearing the ball/area", e.g., \textit{Santander with a header to clear it to safety}. This specific sense is rarely covered in general textual pre-training datasets, hence the failure of \textit{InVeRo}, which relies on BERT, to assign the right sense. The context from the video is, in this case, fundamental to learn a refined representation for such verbs. Studying how to integrate richer context in language models is an interesting avenue for multi-modal representation learning.


\section{Conclusions}

We present \acro, a new dataset that enables the study of knowledge and visual grounding in the sports domain. We show that, in contrast to existing vision+language datasets, models trained on \acro\ require grounding in both language and vision.
We also discuss how external 
information can further improve performance. 
We provide results for four vision+language tasks, including commentary retrieval, frame reordering, video moment retrieval, and commentary generation. 
Our findings on frame reordering and video moment retrieval diagnostics indicate that current pretrained transformer-based models have limited temporal reasoning capabilities.

We consider commentary generation the most challenging task as it requires video understanding on multiple time scales. Our error analysis reveals that the text generated by the examined models matches the style of football commentaries, but it fails to accurately describe the depicted events. We show that generation can be supported by utilising 
a knowledge base.
Designing a multi-modal model able to fuse and integrate all input sources represents an exciting avenue for future research.

\acro\ can also serve as a useful resource for other multi-modal tasks, including:
visual context-aware speech recognition e.g. \cite{ghorbani2021listen} (\acro\ has high quality transcriptions); multi-modal fact-checking (i.e. in a goal event, check if the generated commentary mentions the correct goal scorer); and multi-modal activity recognition. 
Additionally, GOAL represents an interesting benchmark for models that do not require textual annotations but directly solve the task using audio information. Therefore, we consider integrating the audio features provided in GOAL an interesting direction for multi-modal machine learning community.



\clearpage

\bibliographystyle{ACM-Reference-Format}
\bibliography{goal}

\pagebreak

\appendix
\section{Dataset analysis}\label{appendix:dataset_statistics}


\subsection{Dataset statistics}
The resulting dataset has been divided into 3 splits using stratified sampling based on the league label. We completed a first 70-30 sampling obtaining \texttt{train+val} and \texttt{test} splits. Then, we completed a 80-20 sampling obtaining the \texttt{train} and \texttt{val} splits. These splits are the ones that we use as reference for all the tasks that are presented in the \acro{} experimental evaluation. We analysed the collected videos and reported statistics about them in Table~\ref{tab:goal_video_statistics}. The $1107$ videos are extracted from $8$ different leagues such as: the Italian Serie A ($638$ videos), the English Premier League ($167$ videos), UEFA Champions League ($62$ videos), UEFA Europa League ($61$ videos), FA Cup ($34$ videos), Carabao Cup ($18$ videos), EFL Championship ($77$ videos) and EURO 2020 Qualifiers ($50$ videos).


\begin{table*}[]
\centering
\begin{tabular}{@{}lllll@{}}
\toprule
\#videos & Avg. duration (min) & Min. duration (min) & Max. duration (min) & Total duration (min) \\ \midrule
1107     & 3.96                & 1.3                 & 11.6                & 4387.38              \\ \bottomrule
\end{tabular}
\caption{Statistics about the videos contained in the \acro{} dataset. }
\label{tab:goal_video_statistics}
\end{table*}

\subsection{Situated Language Analysis}\label{appendix:situated_language_analysis}

\begin{figure*}[tb]
\centering
    \includegraphics[keepaspectratio,width=0.9\linewidth]{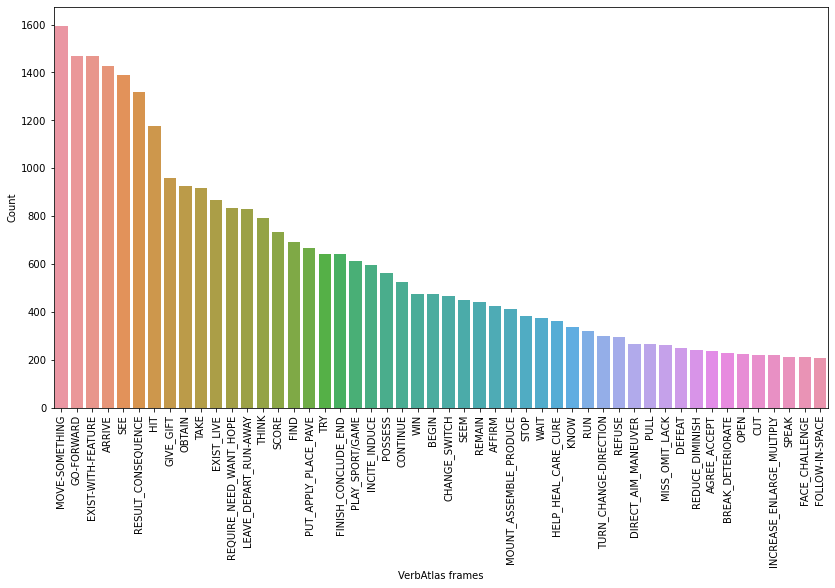}
    \caption{Distribution of the VerbAtlas predicative structures extracted by the \textit{InVeRo} model~\cite{conia2020invero}.}
  \label{fig:goal_verbatlas}
\end{figure*}

As described in the main text, we used the \textit{InVeRo} platform to extract predicative structures from the football commentaries provided in \acro{}. The model is able to annotate predicative structures for several language resources such as VerbAtlas and English Propbank. In this section, we provide the distribution of predicative structures based on VerbAtlas. Similarly to the English Propbank, as shown in Figure~\ref{fig:goal_verbatlas}, the model is able to annotate several motion/action verbs represented by the predicates \textit{MOVE-SOMETHING}, \textit{GO-FORWARD} and \textit{ARRIVE}. Interestingly, predicates specifically associated with the football domain emerge such as \textit{FACE-CHALLENGE}, \textit{SCORE}, \textit{WIN} and \textit{DEFEAT}.

\begin{figure*}[tb]
\centering
    \includegraphics[keepaspectratio,width=0.48\textwidth]{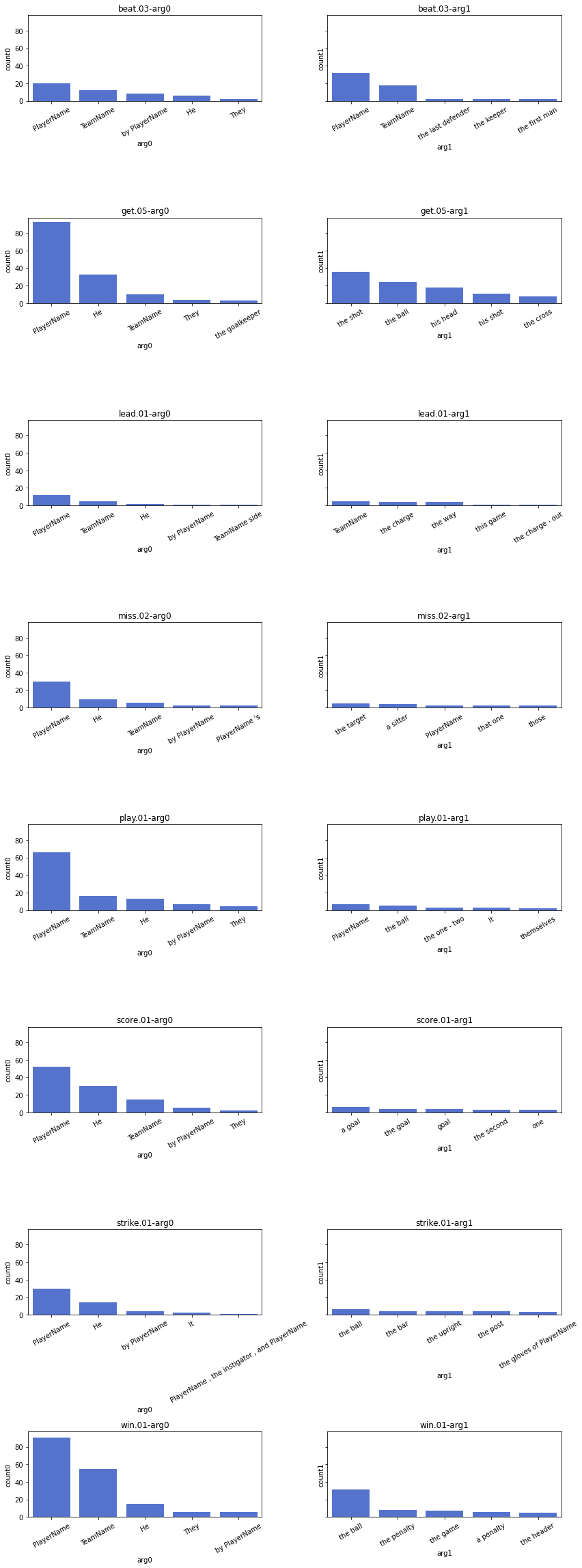}
\caption{Distribution of predicative structures extracted from \acro{} using \textit{InVeRo} model~\cite{conia2020invero}. For the sake of readability, we report only some of the most relevant verbs for the domain of interest.}
  \label{fig:goal_srl_roles}
\end{figure*}

When analysing events is important to understand the role of specific actors and objects involved in them. To understand the role played by entities available in \acro{}, we use semantic roles annotated by the \textit{InVeRo} SRL system. However, annotating the dataset as is, it would involve asking the SRL model to recognise entities (e.g., players, teams, team managers, etc.) that probably has never seen during training. Therefore, we applied a delexicalisation procedure by using our reference Knowledge Base (KB). The knowledge base that we created contains several surface forms for each entity. We implemented an automatic procedure that replaces every occurrence of an entity with its corresponding entity type (e.g., ``'Cristiano Ronaldo' will be replaced by 'PlayerName'"). 
We use \textit{InVeRo} to annotate the delexicalised dataset to extract predicative structures with their semantic roles ARG0 and ARG1. This is because we are interested in how event verbs and their agent/patient (ARG0/1) are relevant to the four tasks, for example, how they help visual language grounding (e.g., “player A passes to player B”). Figure~\ref{fig:goal_srl_roles} presents a summary of the distribution of the ARG0 and ARG1 associated with the most common verbs in the analysed data. An interesting case is definitely the common verb ``to score". The verb is associated with ``the second" which represent a very complex construct to learn. Particularly, three elements of visual grounding are required: 1) understand that somebody scored a goal; 2) who scored the goal; 3) understand that it is the second goal for that player. 

\section{Commentary generation examples}\label{appendix:commentary_gen_examples}

Table \ref{tab:example_captions} presents 5 examples of commentary generate both from BART and HERO, and a scene description to help with cases where that is not possible to infer from the history. In the first example, the target commentary and the HERO picture similar events, one from the point of view of the attacking team and the other from the point of view of the defending team. In the second example, BART fails to understand the half which ended was the first rather than the second, whilst HERO is able to do. In the third example, even if it is not part of the target sentence, the video shows an overlap movement~\footnote{A typical movement which consists of the full back running behind the winger to receive the ball upfront.} of a player in the attacking team, which HERO includes in the commentary. In the fourth example, both models output the same commentary. In the final example, HERO was not able to infer that the ball was over the crossbar and the generated suggest there was a goal.




\section{Experiments setup}\label{appendix:exp_setup}
We design all our experiments using AllenNLP~\cite{gardner2018allennlp} for the sake of better reproducibility. For the HERO model, we use the publicly available code and pretrained checkpoints~\footnote{\url{https://github.com/linjieli222/HERO}} released by \citet{li-etal-2020-hero} and integrate them in the same AllenNLP experimental framework. For all the baselines using Transformer-based architectures, we use the models from HuggingFace Transformers~\cite{wolf2020transformers}.

The hyper-parameters of the baseline models along with the evaluation results on the test set are summarised in Tables~\ref{table:hyperparam-test-frame}-\ref{table:hyperparam-test-select}. We run the experiments on workstations with NVIDIA 2080 GPUs. The maximum training time for a model was around 2 hours. We use the validation data to tune only very important hyper-parameters such as batch size and learning rate. The model reported in the main paper are the ones with the best validation metric. For the KB priming results, the models reported in the paper are trained with history in first position in the context. Different values of KB-source are tested to give an oracle upper bound performance.

\subsection{Frame reordering}

The frame reordering task is intended as a diagnostic task for assessing the capability of the model to pay attention to the position of the visual features. We would like a multi-modal video encoder to be able to understand when its visual features have been shuffled therefore affecting their coherence. For this task we compare the HERO model with several sequence encoders. Specifically, we want to investigate whether the positional encoding used in HERO is able to support the model in completing this task. We compared it with several sequence encoders baseline. Each encoder has a prediction layer on top of its hidden states which is a 2-layers MLP with dropout $0.1$ and $relu$ activation function used to predict the position for the current visual feature. The \texttt{bi-GRU} and \texttt{bi-LSTM} are one layer bidirectional sequence encoders with hidden size $384$. We train the models using Adam~\cite{kingma2014adam} whose parameters are specified in Table~\ref{table:hyperparam-test-frame}.

\subsection{Commentary generation}

For CopyNet, sentences are tokenized into words and embedded by a layer initialized with $300$-dimensional GloVe~\cite{pennington2014glove} embeddings. The embedding layer is fine-tuned during training. The encoder and decoder modules use bi-directional GRUs~\cite{cho2014learning} with $256$ units and $2$ layers. The model is optimized for $100$ epochs using Adam~\cite{kingma2014adam} with learning rate $1e-3$, weight decay $0.01$, and mini-batch size $512$.

\begin{table*}[ht]
\centering
\begin{tabular}{lc|c}
\multicolumn{3}{c}{\textbf{Frame reordering}}\\
\hline
\textbf{Model} & \textbf{LR}  &  \textbf{Accuracy} \\
\hline \hline
MLP & 1e-4 & 10.42 \\
MLP+pos & 1e-4 & 10.42\\
bi-GRU & 1e-3 & 90.00\\
bi-LSTM & 1e-3 & 91.00\\
HERO-flat & 1e-4 & 87.00\\
\hline
\end{tabular}
\hfill
\begin{tabular}{lcc|cc}
\multicolumn{4}{c}{\textbf{Moment retrieval}}\\
\hline
 \textbf{Model}     & \textbf{LR} & \textbf{Soft}  & \textbf{Weighted}       \\ \hline \hline
HERO-flat   & 1e-6  & 70.6 & 2.6 \\
VRoBERTa  & 5e-5  & 32.24          & 2.11          \\ \bottomrule
\end{tabular}
\caption{Hyper-parameter tuning for Learning Rate (LR). Models are trained using the Adam \cite{kingma2014adam} optimizer with weight decay of $0.01$, and batch size $32$.}

\label{table:hyperparam-test-frame}
\end{table*}

\begin{table*}[h]
\centering
\resizebox{\textwidth}{!}{
\begin{tabular}{lcccc|cccc}
\multicolumn{9}{c}{\textbf{Commentary generation}}\\
\hline
\textbf{Model} & \textbf{BS} & \textbf{LR} & \textbf{Top-k} & \textbf{Top-p} & \textbf{BERTScore $\uparrow$} & \textbf{BLEU $\uparrow$} & \textbf{METEOR $\uparrow$} & \textbf{ROUGE-L $\uparrow$}  \\
\hline \hline
BART & 32 & 1e-4 & 100 & 0.95 & 0.84 & 0.79 & 3.80 & 8.79 \\
HERO-flat (Tied Weights) & 32 & 3e-4 & 100 & 0.95  & 0.84 & 0.53 & 3.5 & 6.69 \\
HERO-flat & 32 & 3e-4 & 100 & 0.95 & 0.84 & 0.60 & 3.70 & 7.20 \\
RoBERTa & 32 & 1e-4 & 100 & 0.95  & 0.85 & 0.36 & 2.94 & 5.78 \\
CopyNet & 512 & 1e-3 & 10 & 0.9 & 0.82 & 0.34 & 4.19 & 8.97 \\
\hline
\end{tabular}
}
\caption{Hyper-parameter tuning for batch size (BS) and learning rate (LR). Encoder decoder weights are not tied unless specifically mentioned.}
\label{table:hyperparam-test}
\end{table*}

\begin{table*}[h]
\centering
\small
\footnotesize{
\resizebox{0.85\textwidth}{!}{
\begin{tabular}{lcccc|ccc}
\multicolumn{8}{c}{\textbf{Commentary retrieval}}\\

\hline
\textbf{Model} & \textbf{BS} & \textbf{LR} & \textbf{Modalities} & \textbf{Length} & \textbf{R@1 $\uparrow$} & \textbf{MRR $\uparrow$} & \textbf{Mean $\downarrow$}  \\
\hline \hline
RoBERTa & 16 & 1e-5 & Text & - & 0.37 & 0.60 & 2.31 \\
HERO-flat & 16 & 1e-4 & Vision & - &  0.55 & 0.73 & 1.74 \\
HERO-flat & 16 & 5e-6 & Both & - &  0.57 & 0.75 & 1.71 \\
HERO-flat & 16 & 1e-4 & Both & 50 &  0.58 & 0.75 & 1.67 \\
HERO-flat & 16 & 1e-4 & Both & 80 &  0.59 & 0.76 & 1.65 \\
HERO-flat & 4 & 5e-5 & Both & 100 &  \textbf{0.63} & \textbf{0.79} & \textbf{1.57} \\
\hline
\end{tabular}
}}
\caption{Hyper-parameter tuning for Batch Size (BS), Learning Rate (LR). Evaluation on test set for response selection with 5 candidates. We did a small ablation with different modalities. `Both' in the table signifies using both vision and text modalities. We also experimented with length of the previous 2 textual captions. HERO-flat using both modalities was our best performing model.}
\label{table:hyperparam-test-select}
\end{table*}

\begin{table*}[h]
\scriptsize
    \centering
    \begin{tabulary}{\textwidth}{LLLLL}
        \textbf{History} & \textbf{Target} & \textbf{BART} & \textbf{HERO} & \textbf{Scene} \\
        \hline
        Piatek and Berisha did well. I think a yellow card maybe forthcoming. Long by Donnarumma, flicked on by Piatek, brought down by Çalhanoğlu and Paquetá could be away for number two. Lucas Paquetá, denied by Berisha. Çalhanoğlu with the corner, it's a chance for Léo Duarte. & SPAL can't get it clear and eventually they do with Murgia. & It's off the crossbar
         & They don't want to be effective & Several players from the attacking team attempt to shoot on target \\
         \hline
         Torino right back in it, Baselli, Top corner, Pick that one out. Zaza continues to warm up, Muriel, Oh, that's a lovely ball, Simeone is through here, Oh, he tried a little dink, and it just wasn't enough to take it beyond Sirigu, who, as he was going down, made the save , That is a golden opportunity wasted. Oh, tree minutes have been added on at the end of this half, so time for another goal, possibly, Chance here, It's a great save, again, by Sirigu to the rescue. & It's Fiorentina who get the second period underway. & There goes the final whistle, and Torino run out 2-0 winners here against Sassuolo. & The second half resumes then. & Beginning of the second half \\
         \hline
         Fierce effort from the Italy international. Snapshot from distance. The second half has started. Sampdoria playing from right to left. Defrel, here's Linetty, Quagliarella scoops it in towards Defrel. & Defrel blazes it over. & Oh, what a chance for Quagliarella. & It's really difficult for Defrel and on the overlap from Caputo, he'll have it. & Chance for Defrel after an overlap movement from his team's fullback \\
         \hline
         Milan do have a great recent record against Torino, unbeaten in the last 22 Serie A meetings. Milan defensively strong, No foul there. None whatsoever. Berenguer, Romagnoli in attendance. De Silvestri, another slip for Milan, A player goes down. & The referee points for the penalty. & The referee points to the spot. & The referee points to the spot. & Penalty given by the referee \\
         \hline
         They thought it might have been a foul. Of course, Martin Atkinson will be looking at that incident. One player seem to appeal. It's Maddison. It might open up for him now. & Glances off the crossbar as close as they come. & It's James Maddison. & This will be an attacker's made form and that is 3-0. & Shot over the cross-bar \\
         \hline
    \end{tabulary}
    \caption{Examples of commentaries generated from BART and HERO.}
    \label{tab:example_captions}
\end{table*}

\begin{figure*}[tb]
\centering
    \includegraphics[width=0.9\linewidth]{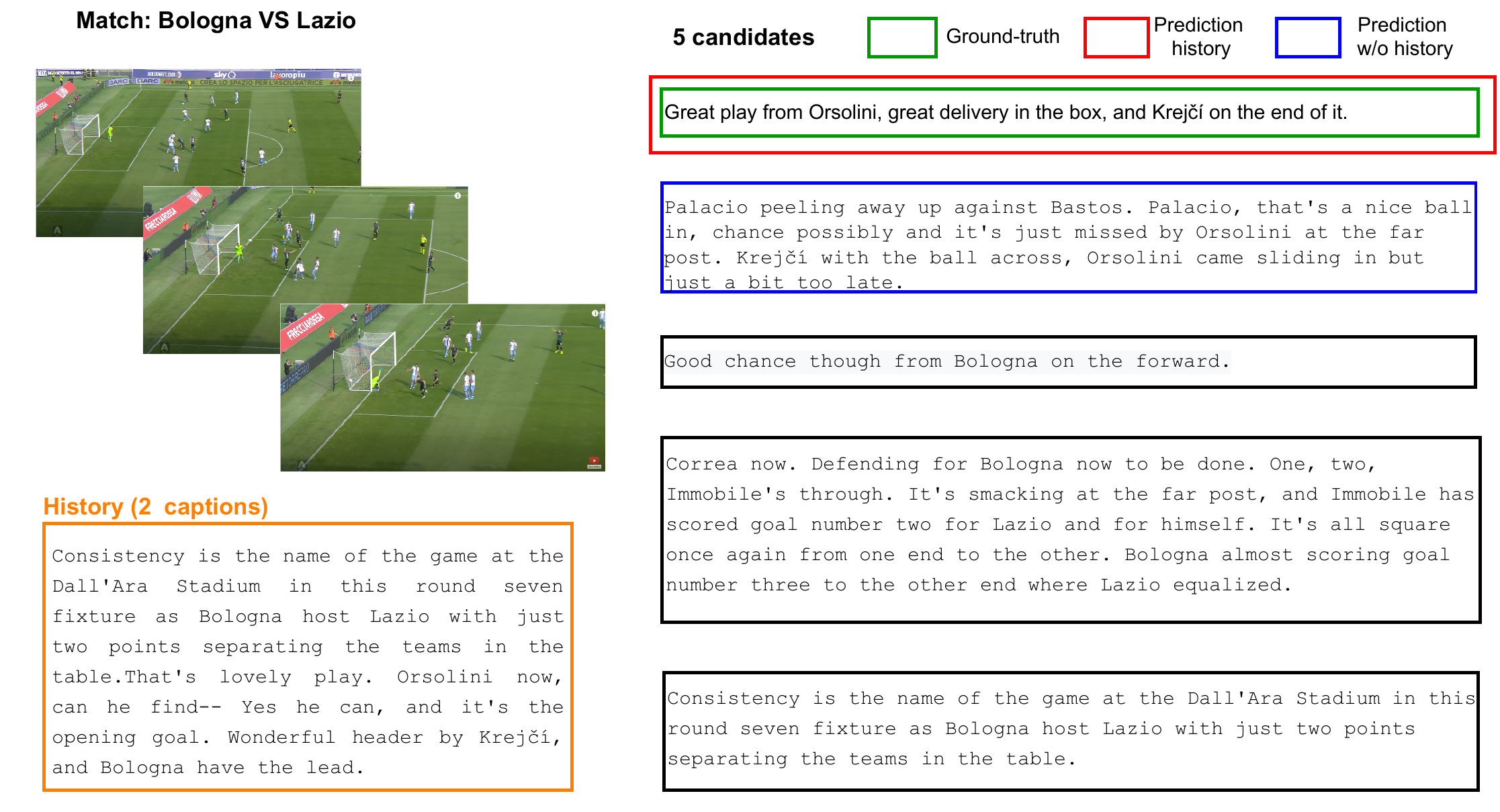}
  \caption{We demonstrate qualitative results on the response retrieval task for the HERO vision-only (no history) and HERO models. For every turn, HERO uses 2 previous captions as history.}
  \label{fig:results}
\end{figure*}

\begin{figure*}[tb]
\centering
    \includegraphics[width=0.8\linewidth]{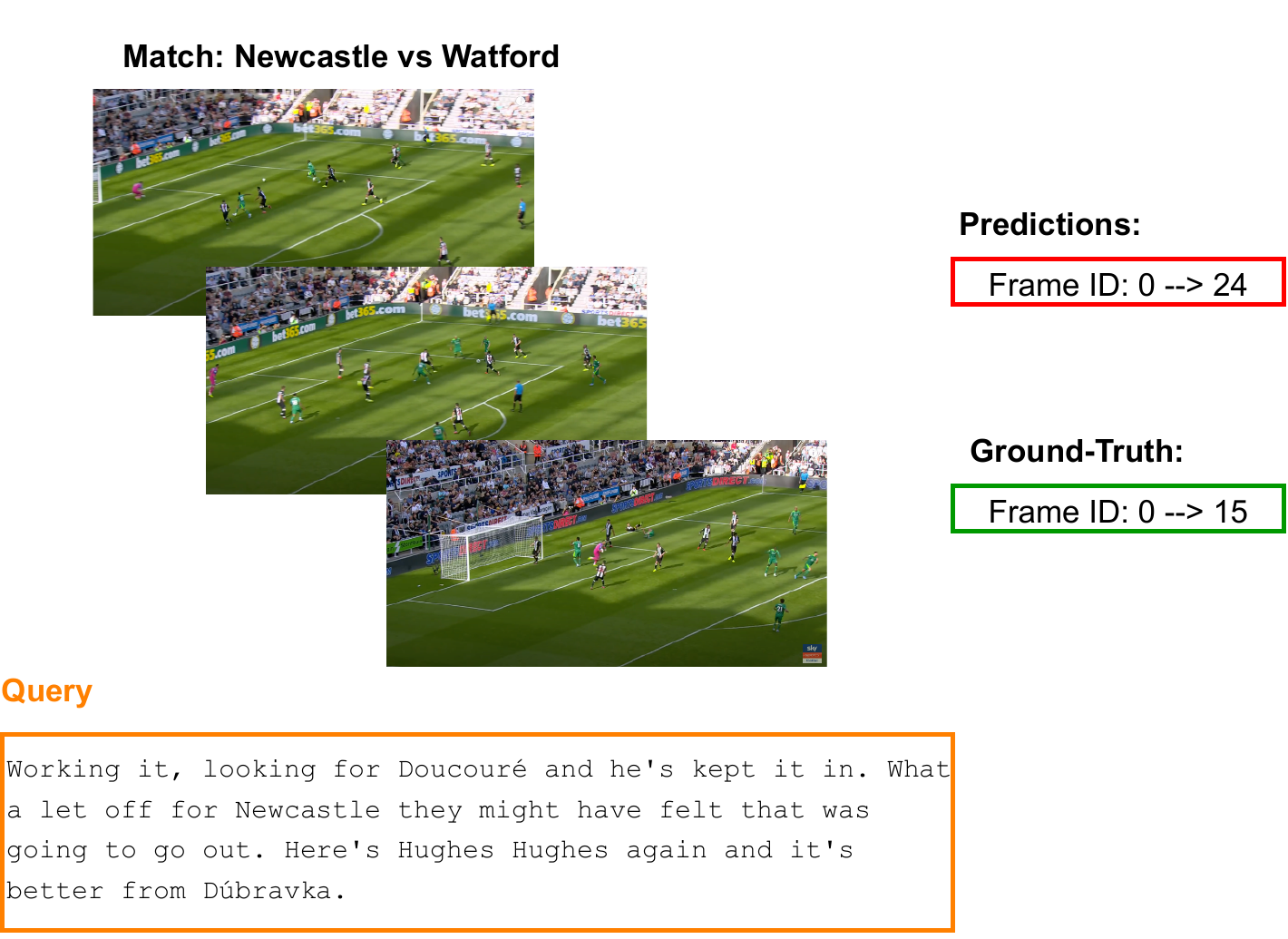}
  \caption{We demonstrate qualitative video moment retrieval results for the HERO model. As discussed in the evaluation section, the model predicts wide ranges which are still meaningful for the task.}
  \label{fig:moment}
\end{figure*}

\section{Responsible NLP Research Checklist}

\subsection{Limitations}
We collected the dataset \acro{} to provide a benchmark for advancing the state of the art in the multimodal modelling community. However, we are aware of several limitations of our work that represent interesting avenue for future work. Most importantly, we collected the dataset for a very specific domain which is football commentaries. This dataset is supposed to be used as a diagnostic benchmark rather than a pretraining dataset. Furthermore, we provide high-quality manual annotations only for 1107 videos which makes GOAL unsuitable for large-scale training.   

\subsection{Potential Risks}
The collected dataset is about football highlights created by official football media providers. Therefore, we consider its content not harmful in any way.

\subsection{License and Terms of Use}

We used open source software and resources that are part of our experimental evaluation. Every mention to such resources has been properly acknowledged in both the main text and appendix. Additionally, to the best of our knowledge, we have used all such software and resources following the corresponding guidelines of fair use.

\subsection{Information about Names and Offensive Content}

We collected a dataset of football commentaries created by reputable media services in the field. Therefore, the videos that are part of \acro{} have a very high quality and do not content any form of offensive content. Additionally, we didn't apply any anonymisation procedure to the data because we're interested in understanding whether the trained models are able to recognise players and their role in the field. This information is essential to generate high quality commentaries. 

\end{document}